%% file: collas2025_conference.tex
\documentclass{article} 
\usepackage{collas2025_conference,times}
\usepackage{easyReview}

\input{math_commands.tex}

\usepackage{booktabs}
\usepackage{multirow}
\usepackage{siunitx}
\usepackage{wrapfig}
\usepackage{subcaption}

\usepackage{hyperref}
\hypersetup{
    colorlinks=true,
    linkcolor=red,
    filecolor=magenta,
    urlcolor=blue,
    citecolor=purple,
    pdftitle={Overleaf Example},
    pdfpagemode=FullScreen,
    }

\usepackage{xcolor}
\definecolor{p_orange}{HTML}{FF7F0e}
\definecolor{p_blue}{HTML}{1f77b4}
\definecolor{p_green}{HTML}{45ad4b}
\definecolor{p_turquoise}{HTML}{40E0D0}
\definecolor{p_yellow}{HTML}{ecc35c}

\title{Forgetting of task-specific knowledge in model merging-based continual learning}

\author{Timm Hess, Gido M van de Ven, Tinne Tuytelaars\\
Department of Electrical Engineering (ESAT), 
KU Leuven, 
Belgium \\
\texttt{\{timmfelix.hess, 
gido.vandeven, 
tinne.tuytelaars\}@esat.kuleuven.be}\\
}

\preprintcopy 

\begin{document}

\maketitle

\begin{abstract}
This paper investigates the linear merging of models in the context of continual learning (CL). Using controlled visual cues in computer vision experiments, we demonstrate that merging largely preserves or enhances shared knowledge, while unshared task-specific knowledge rapidly degrades. We further find that merging models from an incremental training process consistently outperforms merging models trained in parallel.
\end{abstract}

\section{Introduction}
\label{sec:introduction}
Methods that involve averaging the parameters of different models, often termed weight-space ensembling or model merging, have received significant attention in deep learning \citep{yang2024model}. A common application targets improving performance for a single task \citep{izmailov2018averaging, wortsman2022robust}. When multiple (well-performing) model variants are trained starting from a common initialization -- for instance, through multiple fine-tuning runs with different random seeds or different hyperparameter settings -- averaging their weights tends to yield a final model with improved accuracy and robustness over any individual model, without increasing inference cost. This benefit is generally attributed to the geometry of the loss landscape, as models fine-tuned from the same initialization have been found to often reside in the same wide, and relatively flat basin \citep{goodfellow2014qualitatively, frankle2018lottery}. A common interpretation is that averaging the weights of such models produces a solution closer to the center of this basin, smoothing out noise or over-specifications learned by individual training runs, leading to better generalization \citep{izmailov2018averaging}. 

Beyond improving performance for single tasks, model merging has also been explored for integrating knowledge from multiple tasks into a single model \citep{ilharco2022patching, matena2022merging}. This renders model merging a potential mechanism for continual learning (CL), where the goal is to learn multiple tasks sequentially without catastrophically forgetting earlier ones \citep{mccloskey1989catastrophic, parisi2019continual}. With CL in mind, two approaches for merging models that are trained or adapted for different tasks can be distinguished. One option is to merge different states of an incrementally trained model, by averaging its weights after learning one task with its subsequent weights after learning a later task. Another option is to merge separate models trained in parallel on different tasks. Some studies have already shown promising results for model merging in specific CL contexts~\citep{marouf2024weighted, udandarao2024practitioner, kozal2024continual}. Notably, the merging mechanism differs fundamentally from the typical approach to CL, which is to make changes to the loss function to sequentially approximate a joint objective over all observed tasks \citep{hess2023two}. Instead, weight-space merging is typically applied post-hoc, combining independently or sequentially trained models that have been specialized for individual tasks. 
The general conditions under which such post-hoc merging preserves or degrades knowledge across diverse tasks remain poorly understood. To better understand when merging is suitable for CL, we conceptualize the knowledge of a model trained on multiple tasks as comprising both \textit{shared} components (e.g.,~general features from pre-training, or knowledge common across multiple tasks) and \textit{task-specific} components (e.g.,~features or decision boundaries unique to a single task). 
The central question of our study then is: \textit{how do these shared and task-specific knowledge components react during weight merging?} 

In this work, we conduct controlled experiments targeting the computer vision domain. We propose a methodology that allows to instantiate either shared or task-specific knowledge via synthetic visual cues injected directly into the input image space. We empirically show that during linear weight interpolation, shared knowledge tends to be largely preserved or even enhanced, while unshared task-specific knowledge is significantly degraded. These results align with recent research by \citet{zaman-etal-2024-fuse}, who conduct related experiments with large language models. We further compare merging incrementally trained models with merging parallel trained ones. 
Our findings provide insight into the suitability and limitations of weight-space ensembling as a mechanism within various CL scenarios, potentially informing the design of more effective strategies for knowledge accumulation and retention. 

\section{Background}
\label{sec:background}
\textbf{Continual learning.} An important goal of continual learning (CL) is enabling models to learn sequentially from a stream of tasks, 
without catastrophically forgetting previously acquired knowledge~\citep{de2021continual, wang2024comprehensive,ven2025continual}. The dominant approach in CL involves incrementally training a single, evolving model by attempting to constantly approximate a joint learning objective across all encountered tasks \citep{hess2023two}.
In contrast, model merging as a post-hoc method, offers a distinct approach by combining separately adapted model states, rather than modifying the sequential training process itself.

\textbf{Shortcuts.}
To empirically study the interaction of different knowledge types during merging, we draw inspiration from research on shortcut learning. `Shortcuts' refer to spurious or overly simple correlations in data, which the models exploit rather than learning more generalizable features~\citep{geirhos2020shortcut}. By intentionally injecting synthetic visual cues that are correlated with class labels but distinct from core image content, scenarios can be created where models learn to rely on these shortcuts.

For a more comprehensive review of related literature, we refer the reader to Appendix~\ref{app:related_work}.

\section{Methodology}
\label{sec:methodology}
\begin{figure}[t]
    \begin{subfigure}{0.48\textwidth}
        \hspace*{0.5em}
        \includegraphics[width=0.98\textwidth]{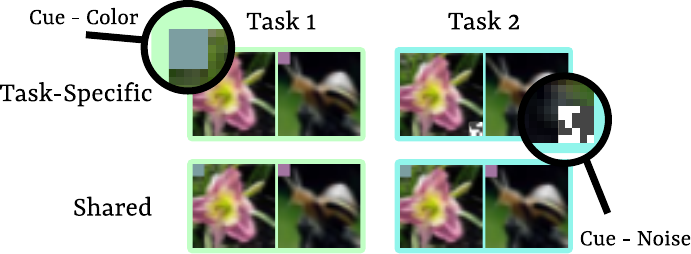}
        \caption{} 
        \label{fig:panel_a}
    \end{subfigure}
    \hfill
    \begin{subfigure}{0.48\textwidth}
        \hspace*{0.5em}
        \includegraphics[width=0.8\textwidth]{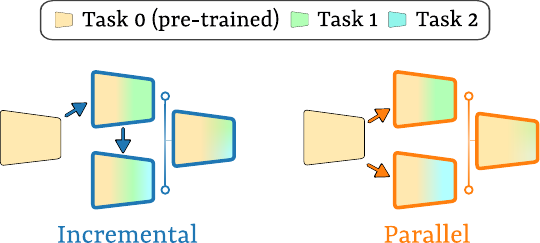}
        \caption{} 
        \label{fig:panel_b}
    \end{subfigure}
    
    \vspace{1em} 
    
    \begin{subfigure}{0.49\textwidth}
        \centering
        \includegraphics[height=3.8cm]{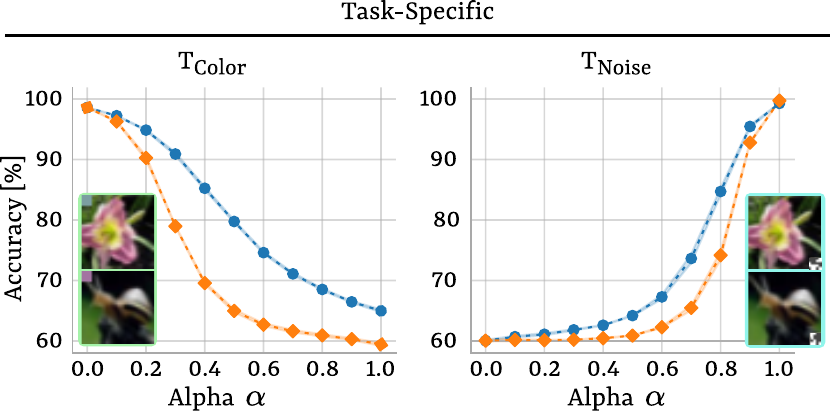}
        \caption{}
        \label{fig:panel_c}
    \end{subfigure}
    \hfill
    \begin{subfigure}{0.24\textwidth}
        \centering
        \includegraphics[height=3.775cm]{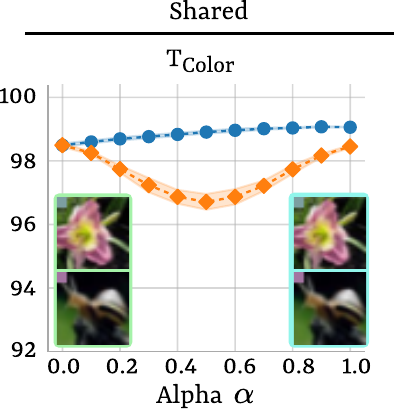}
        \caption{}
        \label{fig:panel_d}
    \end{subfigure}
    \hfill
    \begin{subfigure}{0.24\textwidth}
        \centering
        \includegraphics[height=3.35cm]{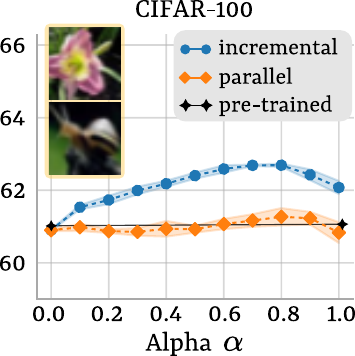}
        \caption{}
        \label{fig:panel_e}
    \end{subfigure}

    \caption{\textbf{Experimental Protocol and Main Results.}
    \textbf{(a)}~Example of shared and task-specific knowledge instantiated with visual cues. The shared knowledge protocol uses the same cue for both tasks, while the task-specific knowledge protocol uses distinct cues.
    \textbf{(b)}~Schematic illustration of `incremental' (\textcolor{p_blue}{blue}) and `parallel' (\textcolor{p_orange}{orange}) training. Both scenarios start from a common pre-trained model (Task 0, \textcolor{p_yellow}{yellow}) and adapt models for subsequent tasks (Task 1, \textcolor{p_green}{green}; Task 2, \textcolor{p_turquoise}{turquoise}) that involve specific visual cues.
    \textbf{(c)}~Accuracy (y-axis) vs.\ interpolation coefficient $\alpha$ (x-axis) for the task-specific knowledge protocol. Performance is evaluated in the presence of the color (left) and noise (right) visual cues, comparing incremental (blue circles) and parallel (orange diamonds) training. The endpoint models of the interpolation are specialized for $T_{\text{Color}}$ ($\alpha=0$) and $T_{\text{Noise}}$ ($\alpha=1$).
    \textbf{(d)}~Accuracy vs.\ $\alpha$ for the shared knowledge protocol, where both endpoint models of the interpolation are specialized for $T_{\text{Color}}$. 
    \textbf{(e)}~Performance on the base dataset (no cues) when interpolating $T_{\text{Color}}$ ($\alpha=0$) and $T_{\text{Noise}}$ ($\alpha=1$) endpoints; the solid black line marks accuracy of the pre-trained model. The plotted lines in the bottom three panels represent the mean accuracy across three independent runs with different random seeds, while the shaded areas indicate the standard error. All evaluations are on the full CIFAR-100 test set.
    \emph{\textbf{These results show that while unshared task-specific knowledge degrades rapidly when models are merged (panel~c), shared knowledge components are largely preserved or even enhanced (panel~d,e). Moreover, merging incrementally trained models consistently leads to better knowledge preservation compared to merging models trained in parallel.}}}
    \label{fig:res_weight_interpolation}
\end{figure}

Our methodology is designed to investigate the fate of shared (common) and unshared (task-specific) knowledge during model merging in computer vision settings.
To create controllable and distinct knowledge components in our models, we augment a base image dataset by superimposing synthetic visual cues onto the input images. Examples are shown in Figure~\ref{fig:panel_a}. A combination of the base dataset with a specific visual cue constitutes a distinct `task' for the model to learn. Interpolating between endpoint models that learned tasks with either different or the same visual cues governs our investigation of task-specific and shared knowledge retention. For all our experiments, CIFAR-100 \citep{krizhevsky2009learning} is used as the base dataset.
The full implementation details are presented in Appendix~\ref{app:hyperparams}.

\textbf{Shared pre-trained initialization.}
All model trainings begin from a common base model, $\theta_{\text{PT}}$, which has been pre-trained on the base dataset. The primary purpose of pre-training is to ensure that all endpoint models share a significant initial learning trajectory, a condition motivated by linear mode connectivity research, which suggests it facilitates meaningful weight-space merging without requiring complex re-parameterization techniques~\citep{goodfellow2014qualitatively, frankle2018lottery, ainsworth2022git}. In addition, the pre-training on the base dataset (without cues) yields a broad set of general features, which allows us to evaluate the preservation of this foundational `shared knowledge' as an independent measure alongside our visual cue-specific evaluations, which we detail next.

\textbf{Visual cues.}
We define a visual cue as an $N \times N$ pixel patch superimposed onto an image at a specific location, providing a simple visual pattern that is consistently correlated with the sample's ground-truth class label, irrespective of the underlying image content.
To create two different types of shortcuts, we define two families of cues: 
(1)~Colored patches, consisting of pixel patches of solid color~(the \emph{color cue}, denoted as~$\mathcal{C}_{\text{color}}$). We utilize the HSV color space, setting Saturation~(S) and  Value~(V) to $1.0$, and distributing Hue~(H) evenly across classes. 
(2)~Grayscale noise patches, consisting of pixel patches containing grayscale noise patterns~(the \emph{noise cue}, $\mathcal{C}_{\text{noise}}$). For each class, a unique 
pixel noise pattern is generated by sampling each pixel's intensity independently and uniformly from the discrete set $\{0, 255\}$. 
For both families of cues, we use a patch size of~$N=5$. During training, a cue is superimposed on each image with a probability $p_{\mathcal{C}} = 0.5$ to incentivize models to learn both the cue and the general image features.

\textbf{Shared vs.\ task-specific knowledge.}
We use the visual cues to construct two knowledge protocols that are at the core of our experimental setups.
(1)~The \textit{unshared, task-specific knowledge protocol} consists of two tasks constructed using distinct visual cues: in one task the color cue is added to the base dataset~(denoted as~$T_{\text{color}}$), and in the other task the noise cue is added~($T_{\text{noise}}$). To minimize information transfer, the cues are placed at non-overlapping positions (top-left and bottom-right).
(2)~The \textit{shared knowledge protocol} consists of two tasks with the same visual cue: in both tasks, the color cue is added to the base dataset at the same position (top-left).
This controlled use of cues allows for a more explicit separation of shared versus unshared specific knowledge compared to typical CL benchmarks based on semantic splits, where disentangling preserved from re-discovered knowledge can be challenging~\citep{hess2023knowledge}.

\textbf{Incremental vs.\ parallel training.}
Starting from the pre-trained weights~$\theta_{\text{PT}}$, we train models on one of the knowledge protocols to generate \emph{endpoint models} (i.e., models that are trained or adapted for a task with a particular visual cue) for our merging analysis.
As illustrated in Figure~\ref{fig:panel_b}, we compare two distinct training scenarios. With incremental training, which represents continual learning without specific forgetting mitigating, factors like catastrophic forgetting or knowledge transfer between the sequential training stages can influence the characteristics of the endpoint models. In contrast, parallel training serves as a controlled baseline where both endpoint models are trained independently from the identical pre-trained state~$\theta_{\text{PT}}$. 
This setup allows for a direct analysis of merging effects with and without sequential dependencies.

\textbf{Evaluation protocol.}
We evaluate all models (pre-trained, endpoints, and interpolated) using classification accuracy on the held-out test set. To probe shared and task-specific knowledge induced by visual cues, we evaluate performance on the test set with either the color cue ($\mathcal{C}_{\text{color}}$) or the noise cue ($\mathcal{C}_{\text{noise}}$) deterministically ($p_C=1.0$) applied. To measure the preservation of general shared knowledge, we evaluate performance on the original test set without any visual cues ($p_C = 0.0$) applied.
Interpolated models $\bar{\theta}(\alpha) = \alpha \theta_{T_1} + (1-\alpha) \theta_{T_2}$ are generated via linear interpolation of appropriate task-specific endpoints, with $\alpha \in [0, 1]$ the interpolation coefficient.


\section{Results}
\label{sec:results}
Here we present the results of the merging experiments, with the endpoint models generated via either parallel or incremental training and according to either the shared or task-specific knowledge protocol. Key trends are summarized in Figure~\ref{fig:res_weight_interpolation}, with full numerical details presented in Table~\ref{tab:accuracy} in Appendix~\ref{app:result_details}. 

\textbf{Endpoint models successfully learn specific cues while retaining general knowledge.}
Before investigating model interpolation, we confirm that the endpoint models behave as expected. Models adapted to a specific visual cue demonstrate high performance when tested on that cue, with cue-specific accuracies consistently above $98.5\%$ (Table~\ref{tab:accuracy}), while performance on the shared general knowledge (CIFAR-100 without cues) remains close to the pre-trained baseline of $61.3\%$. An exception is the endpoint model incrementally trained on both cues, whose performance on CIFAR-100 without cues surpasses that of the pre-trained baseline, indicating positive transfer from the sequential training process.

\textbf{Unshared specific knowledge degrades while shared knowledge is preserved or enhanced.}
A central finding of our work is the starkly different effect of linear interpolation on different knowledge types. As shown in Figure~\ref{fig:panel_c}, unshared task-specific knowledge rapidly degrades, i.e. sensitivity to a specific cue decays sharply as the interpolation moves towards the other endpoint, a transition that is particularly abrupt in the parallel setting. Conversely, \textit{shared knowledge} is robust to interpolation. As shown in Figure~\ref{fig:panel_d}, when merging models that were trained on tasks with the same cue ($T_{\text{Color}}$), performance is maintained well across the interpolation path. Furthermore, as shown in Figure~\ref{fig:panel_e}, performance on the common CIFAR-100 task remains stable (in the parallel scenario) or is even enhanced (in the incremental scenario), peaking at an accuracy of $62.69\%$, which is above both endpoints.

\textbf{Merging incrementally trained models preserves knowledge better than merging parallel-trained ones.}
We also find a key distinction between training scenarios, as merging models from an incremental training process consistently outperforms merging models trained in parallel (Figure~\ref{fig:panel_c}~to~\ref{fig:panel_e}). In particular, when merging two models that are adapted for the same specific cue ($T_{\text{Color}}$), the incremental scenario maintains high performance across the entire interpolation path, whereas the parallel scenario exhibits a slight concave dip in accuracy around the midpoint (Figure~\ref{fig:panel_d}). Furthermore, for general shared knowledge (CIFAR-100 without cues, Figure~\ref{fig:panel_e}), the incremental scenario produces a notable synergistic effect: accuracy rises to a peak of $62.69\%$, surpassing both endpoints and the original pre-trained model. In contrast, the parallel scenario's performance remains largely flat, showing no significant benefit from interpolation. 
This finding, supported by related work of~\citep{marouf2024weighted}, is particularly relevant for CL, as it indicates that merging states from a single, evolving model can be more advantageous than merging independently trained specialists.


To assess the generality of these observations, in Appendix~\ref{app:add_results} we conduct additional experiments under varied conditions. We analyze the reverse order of cue adaptations (i.e.,~first $T_{\text{Noise}}$, then $T_{\text{Color}}$), and we explore a `chunking'-setup \citep{lee2023chunking}, where the pre-trained model's feature base is weaker and parts of the previously shared common knowledge of CIFAR-100 are divided over the different tasks. 
The results for both configurations corroborate the distinct behaviors of shared and task-specific knowledge during interpolation that we observe in our main experiments.

\section{Discussion}
\label{sec:discussion}
Our experiments demonstrate that linear model merging distinctly impacts different knowledge types. On one hand, shared knowledge is largely preserved and can be enhanced, aligning with earlier findings of consolidating commonalities within shared loss basins~\citep{izmailov2018averaging}. In stark contrast, unshared task-specific knowledge rapidly degrades upon interpolation due to interference between divergent parameter adaptations, a finding consistent with prior work on large language models~\citep{zaman-etal-2024-fuse}. Another important finding of our work is that merging incrementally trained models yields better knowledge consolidation than merging parallel-trained ones, suggesting 
that the process of sequential adaptation, even without explicit CL mechanisms, guides models along trajectories that are more amenable to beneficial merging. 
However, also when incrementally trained models are merged, unshared task-specific knowledge is still rapidly degraded.
For CL, this implies that while naive merging might strengthen the generality of shared features, it risks catastrophic forgetting of unique past task knowledge. This raises critical questions about the `utility' of what is preserved versus forgotten, an issue orthogonal to the stability-plasticity dilemma. Therefore, the suitability of simple merging for CL appears limited to scenarios prioritizing general shared capabilities or involving highly similar tasks, unless model scale, endpoint training, or merging techniques are specifically tailored to mitigate these trade-offs. We provide an extended discussion in Appendix~\ref{app:extended_discussion}.

\section{Limitations and Future Work}
\label{sec:future_work}
Our study offers initial insights into how linear merging of models differentially affects shared and task-specific knowledge in the vision domain. 
While our controlled setting based on visual cues enabled clear distinctions, several avenues warrant further exploration to better understand the applicability of merging in continual learning.
One is to investigate whether the observed dynamics of shared versus specific knowledge scale to larger, more diverse architectures (e.g.,~vision transformers) and more complex, real-world task sequences. While merging benefits are known for large models
, the precise nature of knowledge interaction as identified here needs validation at scale.
Another compelling direction for future work is to explore methods for achieving more explicit internal disentanglement of general versus task-specific knowledge within models during their (continuous) training. 


\subsubsection*{Acknowledgments}
This work has been supported by funding from the European Union under the Horizon 2020 research and innovation program (ERC project KeepOnLearning, grant agreement No.~101021347) and by a senior postdoctoral fellowship from the Research Foundation -- Flanders (FWO) under grant number 1266823N.

\bibliography{collas2025_conference}
\bibliographystyle{collas2025_conference}

\newpage

\appendix
\section*{Appendix}
The following Appendix provides supplementary material to accompany the main paper. It offers:
\begin{itemize}
    \item[\ref{app:hyperparams}] Further details on our experimental setup 
    \item[\ref{app:related_work}] An extended discussion of related literature
    \item[\ref{app:result_details}] Comprehensive numerical results for the primary experiments
    \item[\ref{app:add_results}] Additional experimental validations under varied conditions
    \item[\ref{app:extended_discussion}] An extended discussion of our results
\end{itemize}

\section{Experimental Setup Details}
\label{app:hyperparams}
This section provides further details on the experimental setup and hyperparameters used throughout all training phases (pre-training and subsequent cue adaptations) described in Section~\ref{sec:methodology} and Appendix~\ref{app:add_results}, unless otherwise specified.

\textbf{Model Architecture and Optimization.}
The core architecture is a slim ResNet-18 model~\citep{lopez2017gradient}. In this model, batch normalization layers were replaced with group normalization layers using a single group to approximate layer normalization, thereby avoiding potential confounding effects from batch normalization's running statistics parameters~\citep{kozal2024continual}.
All training stages (pre-training and subsequent cue adaptations) utilize a cross-entropy loss function and the same optimization settings. Optimization is performed using Stochastic Gradient Descent (SGD) with a momentum of $0.9$ and weight decay of $5 \times 10^{-4}$. Each training stage proceeds for $50$ epochs. A linear learning rate warm-up is applied during the first $5\%$ of these epochs, increasing the learning rate from $0.004$ to $0.1$. Subsequently, a cosine annealing learning rate schedule decays the learning rate from $0.1$ down to a minimum of $1 \times 10^{-5}$ over the remaining epochs. The mini-batch size is $128$. All cue adaptation experiments are repeated three times with distinct random seeds, and reported as mean and standard error. Unless explicitly specified, pre-training is not subjected to different random seeds, but the same pre-trained weights are used as initialization for all runs.

\textbf{Data Augmentation and Visual Cue Application.}
Standard data augmentations for CIFAR-100 are employed during all training stages: random horizontal flips (with a probability of $0.5$) and random crops (image size $32\times32$ pixels, with padding of four pixels).
Visual cues, when applied during the cue adaptation stages, are superimposed onto the images \textit{after} these standard augmentations. For the visual cues (both $\mathcal{C}_{\text{color}}$ and $\mathcal{C}_{\text{noise}}$), we use a patch size of $5 \times 5$ pixels. When distinct cues are used to instantiate unshared task-specific knowledge (e.g., $T_{\text{Color}}$ vs. $T_{\text{Noise}}$ experiments), they are placed at fixed, non-overlapping, and distinct positions: the color patch in the top-left corner and the noise patch in the bottom-right corner of the image.

The code to reproduce our experiments is available at: \url{https://github.com/TimmHess/MergeForget}.



\section{Related Work}
\label{app:related_work}
\textbf{Model Merging and Continual Learning.}
Combining the parameters of multiple neural networks, often referred to as model merging or weight-space ensembling, has gained significant interest \citep{li2023deep, yang2024model}. In single-task settings, averaging model weights, particularly those fine-tuned from a common pre-trained initialization, often improves performance and robustness (`model soups' \citep{wortsman2022model}), potentially finding solutions superior to any individual model \citep{izmailov2018averaging, wortsman2022robust}. 
Techniques vary from simple averaging \citep{wortsman2022model} and interpolation \citep{ilharco2022patching} to more sophisticated methods like task arithmetic \citep{ilharco2022editing, ortiz2023task}, Fisher-weighted averaging \citep{matena2022merging}, and interference resolution strategies \citep{yadav2023ties, marczak2024magmax}. 
The potential of merging extends to continual learning~(CL), where models must learn sequentially without catastrophic forgetting \citep{mccloskey1989catastrophic, parisi2019continual}. 
Commonly, CL methods utilize regularization, replay, architectural, or a combination of these mechanisms, and focus on preserving performance on past tasks during sequential training~\citep{de2021continual, wang2024comprehensive}. Model merging offers a different perspective, as it involves the post-hoc combination of models trained independently or sequentially. Indeed, simple averaging has shown promise in the CL `chunking' setting \citep{lee2023chunking}, was combined with replay \citep{marouf2024weighted}, and has been explored in combination with other techniques for continuous adaptation, sometimes incorporating Fisher information or exponential moving average-like updates sequentially \citep{udandarao2024practitioner, dziadzio2024merge}. 
However, merging models adapted to different tasks introduces interference \citep{yadav2023ties, marczak2024magmax}, a key challenge shared with the stability-plasticity dilemma inherent to CL. 
Building on work distinguishing shared and task-specific knowledge \citep{zaman-etal-2024-fuse}, which found that merging can be used to selectively forget task-specific information in large language models~(LLMs), our work investigates these dynamics in the vision domain to improve our understanding of the suitability of model merging for CL. 

\textbf{Loss Landscapes and Mode Connectivity.}
The effectiveness of model merging, particularly linear interpolation of weights, is closely tied to the concept of mode connectivity \citep{garipov2018loss, draxler2018essentially}. Research suggests that minima found via fine-tuning from a common pre-trained initialization often lie within the same loss basin and can be connected by linear paths of low loss \citep{frankle2018lottery, neyshabur2020being}. This provides a geometric explanation for why averaging fine-tuned models can be successful \citep{wortsman2022model}. 
The situation is less clear when merging models that were adapted to different tasks or data distributions. In that case interference might occur \citep{yadav2023ties, mirzadeh2020linear}.
Our work uses interpolation across models trained with different task-specific cues to implicitly probe these geometric properties. It reveals the nuance that depending on whether knowledge is shared between models or not, merging either preserves that knowledge or leads to interference. 
This finding aligns with the intuition that incompatible representations (likely residing in different basins or requiring non-linear paths) are harder to merge effectively via simple averaging. 

\textbf{Merging Shared and Task-Specific Knowledge -- Fuse-to-Forget.}
While model merging is often explored for its potential to aggregate capabilities in a model, the work of \citet{zaman-etal-2024-fuse} investigated the inverse potential: using merging as a mechanism to selectively \textit{forget} specific, potentially undesirable knowledge components. They investigated this using LLMs. Motivated by applications such as reducing societal biases learned during pre-training or mitigating privacy risks by erasing memorized training examples, the authors postulate that merging might preferentially discard non-shared information.
To analyze this phenomenon systematically, they proposed a framework that distinguishes between \textit{shared knowledge} (common among the models being merged) and \textit{task-specific knowledge} (unique to individual models).
Using this framework, \citet{zaman-etal-2024-fuse} evaluated the effect of merging LLMs fine-tuned with specific, targeted interventions, namely associating disconnected token pairs. 
In their work, this property of forgetting knowledge specific to a particular model is framed as \textit{desirable}, e.g.\ for removing unwanted biases. However, generally speaking, we argue that this property is ambivalent, as forgetting specific task capabilities might be undesirable in other contexts. Our work adapts the concept of shared and task-specific knowledge to the vision domain using a different method to instantiate task-specific knowledge.

\textbf{Cues, Shortcuts, Confounders.}
To create distinct task-specific knowledge components in a controlled manner in the vision domain, we draw inspiration from the literature on shortcut learning \citep{geirhos2020shortcut}. This research area studies how deep neural network models are prone to exploiting spurious correlations instead of learning robust features. By injecting synthetic visual cues that are easy to learn and correlated with the class label, but distinct from the core image content, we encourage models to specialize on these particular `shortcuts'. This setup allows for a cleaner separation between shared (same shortcut in each task) and task-specific (different shortcut in each task) knowledge compared to using standard datasets with semantic splits, where the nature of shared vs.\ specific features can be ambiguous \citep{hess2023knowledge}.
It provides a visual analog to the token-pair association used by \citet{zaman-etal-2024-fuse} and allows us to study how merging interacts with models reliant on different, easily controlled, task-specific strategies. This controlled approach is also related to \citet{busch2024truth}, who study the impact of confounders to CL, but we focus specifically on the interaction of shared and task-specific knowledge when using merging-like approaches. 

\section{Full Numerical Results}
This section presents the comprehensive numerical data of the main experiments discussed in Section~\ref{sec:results} and visualized in Figure~\ref{fig:res_weight_interpolation}. Table~\ref{tab:accuracy} details the mean classification accuracies and standard errors (across three runs) for all evaluated conditions: endpoint models ($\alpha=0$ and $\alpha=1$) and interpolated models ($\alpha \in (0,1)$), trained in either the task-specific knowledge protocol ($T_{\text{Color}}$ $\rightarrow$ $T_{\text{Noise}}$) or the shared knowledge protocol ($T_{\text{Color}}$ $\rightarrow$ $T_{\text{Color}}$), for both parallel and incremental training scenarios, and evaluated with or without the visual cues used during training.
\label{app:result_details}

\input{gfx/results_table_acc_V2}

\section{Additional Experiments}
This section describes experiments designed to further probe and validate the observations presented in the main paper. These include experiments utilizing a `chunked' training approach, and an analysis with the reversed cue adaptation order.

\label{app:add_results}
\subsection{Model Merging in the Chunking-Setup}
To further investigate the fate of shared and task-specific knowledge under model merging, we investigate a condition where the pre-trained feature base is weaker and where parts of the previously shared and pre-trained knowledge are divided over the different tasks. 

\textbf{Experimental Setup.}
Following \citet{lee2023chunking}, we divide the CIFAR-100 training dataset randomly into three chunks of approximately equal size. We checked that each chunk contains examples from all classes.
The pre-training phase now consists of training only on the data of chunk~1. As before, this model, denoted $\theta_{\text{PT-chunk}}$, serves as the common starting point for subsequent adaptations.

The methodology is analogous to our experiments in the main text. We adapt models using visual cues, employing both parallel and incremental scenarios originating from $\theta_{\text{PT-chunk}}$.
For parallel adaptation, one model branch is trained on chunk~2 augmented with the $\mathcal{C}_{\text{color}}$ cue, and another independent branch is trained on chunk~3 augmented with the $\mathcal{C}_{\text{noise}}$ cue. For incremental adaptation, the model is first trained on chunk~2 with $\mathcal{C}_{\text{color}}$, and then training continues using chunk~3 augmented with $\mathcal{C}_{\text{noise}}$.
The visual cues, training objectives, probabilistic cue application ($p_{\mathcal{C}}=0.5$), and optimization hyperparameters remain the same (see Appendix~\ref{app:hyperparams}). Note that in contrast to the experiments in the main text, here, the pre-training is subject to the random seeds of the individual runs.

Finally, linear interpolation is performed between the endpoint models obtained from the cue-specific adaptation phase for both parallel and incremental scenarios. All models, and interpolated models are evaluated on the full CIFAR-100 test set, using the three conditions defined in Section~\ref{sec:methodology}: color cue specific knowledge, noise cue specific knowledge, and shared general knowledge (CIFAR-100 with no cues).

\begin{figure}[htbp!]
    \centering
    \hfill
    \begin{subfigure}{0.48\textwidth}
        \includegraphics[height=3.8cm]{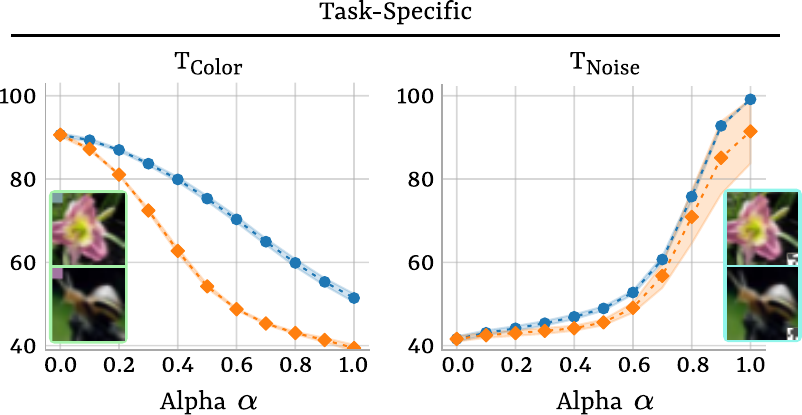}
        \caption{}
        \label{apx_fig:panel_a}
    \end{subfigure}
    \hfill
    \begin{subfigure}{0.24\textwidth}
        \includegraphics[height=3.775cm]{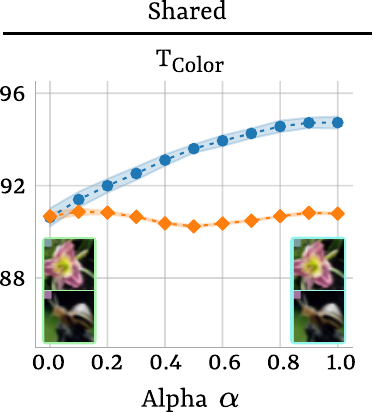}
        \caption{}
        \label{apx_fig:panel_b}
    \end{subfigure}
    \hfill
    \begin{subfigure}{0.24\textwidth}
        \includegraphics[height=3.35cm]{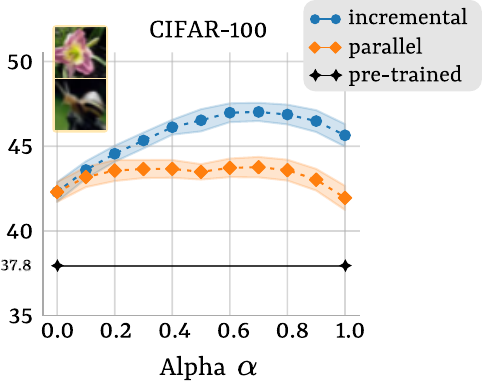}
        \caption{}
        \label{apx_fig:panel_c}
    \end{subfigure}
    \hfill
    \caption{\textbf{Weight-Interpolation Results in the CIFAR-100 `Chunking' Setup.}
    \textbf{(a)}~Accuracy (y-axis) vs.\ interpolation coefficient $\alpha$ (x-axis) for the task-specific knowledge protocol in the chunking setup. Performance is evaluated on the color (left) and noise (right) visual cues, comparing incremental (blue circles) and parallel (orange diamonds) training. The endpoint models are specialized on chunk~2 with $T_{\text{Color}}$ ($\alpha=0$) and chunk~3 with $T_{\text{Noise}}$ ($\alpha=1$), both originating from a base model pre-trained on chunk~1.
    \textbf{(b)}~Accuracy vs.\ $\alpha$ for the shared knowledge protocol, where the endpoint models of the interpolation are specialized for chunk 2 with $T_{\text{Color}}$ ($\alpha=0$) and chunk 3 with $T_{\text{Color}}$ ($\alpha=1$). 
    \textbf{(c)}~Performance on the base CIFAR-100 dataset (no cues) when interpolating the same endpoints as in~(a). The solid black diamond marks the accuracy of the model pre-trained only on chunk~1.
    In all panels, the plotted lines represent the mean accuracy across three independent runs with different random seeds, while the shaded areas indicate the standard error. All evaluations are on the full CIFAR-100 test set.
    \emph{\textbf{These results corroborate our main findings: also when general knowledge is learned distributively across data chunks, merging leads to the rapid degradation of unshared task-specific knowledge (panel~a) while preserving and enhancing the consolidated shared knowledge (panel~b,c).}}}
    \label{fig:res_weight_interpolation_chunking}
\end{figure}

\input{gfx/chunking_result_table}

\textbf{Results and Observations.}
The interpolation results for this chunking experiment are presented in Figure~\ref{fig:res_weight_interpolation_chunking}.
The overall trends we observe are largely consistent with our main findings, though with some nuances reflecting the modified pre-training.

\textit{Endpoint Performance:} The $\theta_{\text{PT-chunk}}$ model (black diamond in the CIFAR-100 panel of Figure~\ref{fig:res_weight_interpolation_chunking}) establishes the baseline shared knowledge accuracy after seeing only chunk~1. Models subsequently adapted to specific cues on new data chunks (chunks~2 and~3) achieve high accuracy when tested with their respective cues on the full test set, indicating successful learning of these specific visual features. Performance of these endpoints on the general CIFAR-100 task (no cues) reflects both the retained knowledge from chunk~1 and any generalizable features learned from chunks~2 and~3.

\textit{Unshared Task-Specific Knowledge:} Sensitivity to the distinct $T_{\text{Color}}$ and $T_{\text{Noise}}$ cues shows rapid degradation during interpolation between the two cue-specialized models, for both parallel and incremental scenarios. This aligns with the main experiments, suggesting that unshared specific knowledge is poorly preserved by merging.

\textit{Shared Task Knowledge:} In the incremental scenario, sequentially training with the same color cue yields forward transfer, strengthening the model's ability to use the cue in the context of new data. Consequently, a monotonic improvement in performance along the interpolation path (Figure~\ref{apx_fig:panel_b}) is observed. Here, merging does not benefit, but visibly reverts the accumulation of knowledge with respect to the visual cue. 
Results for the parallel training scenario, without any forward transfer, show the same trends as our experiment in the main text. A slight concave dip in accuracy at the merging midpoint suggests that specializing on the same cue in the context of different data leads to parametrically distinct solutions that are not perfectly compatible under linear averaging.

\textit{Shared General Knowledge (CIFAR-100):} Performance on the full CIFAR-100 test set (no cues) during interpolation shows improvements for both endpoints, and further substantial gains from interpolation. 
It contrasts the above results for shared specific task knowledge and re-emphasizes a key finding of the main: while merging may revert the learning of a \emph{specific} shared skill in a sequential setting, it can simultaneously consolidate and improve the more distributed, \emph{genera}l shared knowledge acquired across the same sequence.

This experiment and its findings reinforce our main observations. Also when the general features relevant to the CIFAR-100 task are learned in a distributed manner across different data chunks (each potentially paired with a specific cue during adaptation), linear model merging demonstrates a capacity to consolidate and enhance this broadly applicable knowledge. Concurrently, it continues to struggle with reconciling unshared, task-specific cue knowledge introduced within individual chunks. The consistency of these shared-versus-specific dynamics, whether general knowledge is established via comprehensive pre-training or learned incrementally across data chunks, underscores their robustness.

\subsection{Interpolation Results -- Reverse Order of Cue Adaptation}
To further assess the robustness of our main findings regarding the interaction of shared and task-specific knowledge, we conducted an additional set of experiments where the order of cue adaptation in the incremental scenario was reversed compared to that primarily presented in Section~\ref{sec:methodology}.

\textbf{Experimental Setup Modification.}
The core methodology for pre-training, visual cue design ($\mathcal{C}_{\text{color}}$ and $\mathcal{C}_{\text{noise}}$), optimization, and evaluation remains identical to our main experiments (details in Appendix~\ref{app:hyperparams}). The key difference lies in the reversed order of the cues in the task-specific adaptation phase.

\begin{figure}[htbp!]
    \centering
    \begin{subfigure}{0.48\textwidth}
        \includegraphics[height=3.8cm]{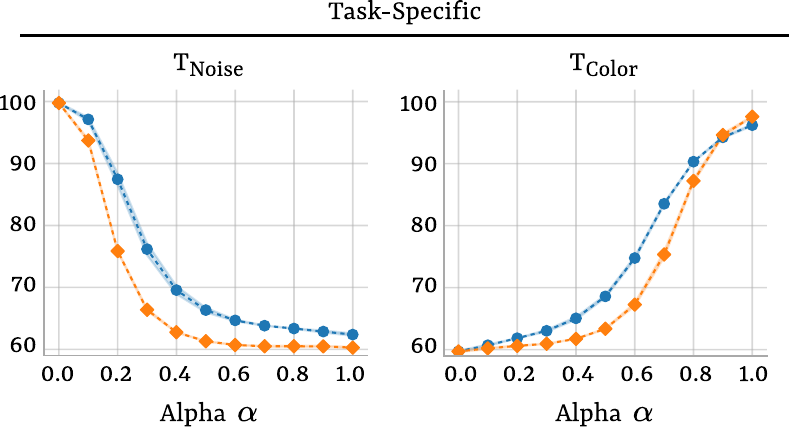}
        \caption{}
        \label{apx_fig_reverse:panel_a}
    \end{subfigure}
    \begin{subfigure}{0.24\textwidth}
        \includegraphics[height=3.35cm]{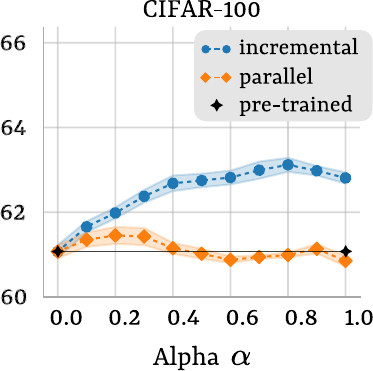}
        \caption{}
        \label{apx_fig_reverse:panel_b}
    \end{subfigure}
    \hfill
    \caption{\textbf{Weight-Interpolation Results for Reverse Order Cue Adaptation.}
    Accuracy (y-axis) vs. interpolation coefficient $\alpha$ (x-axis). The $\alpha=0$ endpoint is specialized on $T_{\text{Noise}}$ and the $\alpha=1$ endpoint on $T_{\text{Color}}$. Results compare `incremental' (blue circles) and `parallel' (orange diamonds) training. 
    \textbf{(a)} The task-specific panels ($T_{\text{Noise}}$, $T_{\text{Color}}$) show the performance on the respective cue-specific test sets.
    \textbf{(b)} The shared CIFAR-100 panel shows the performance without cues.
    In both panels, the plotted lines represent the mean accuracy across three independent runs with different random seeds, while the shaded areas indicate the standard error. All evaluations are on the full CIFAR-100 test set.}
    \label{fig:res_weight_interpolation_reverse_order}
\end{figure}

\textbf{Results and Observations.}
The interpolation results for this reverse order experiment are presented in Figure~\ref{fig:res_weight_interpolation_reverse_order} and Table~\ref{app:tab_cues_reverse_accuracy}. The observed trends are qualitatively consistent with those reported in our main results (Section~\ref{sec:results}. Unshared task-specific knowledge (sensitivity to the initial $T_{\text{Noise}}$ cue at $\alpha=0$ or the final $T_{\text{Color}}$ cue at $\alpha=1$) rapidly degrades when interpolating towards the opposing endpoint. Performance on the shared CIFAR-100 task (no cues) again shows that the incremental adaptation scenario can lead to an enhancement of this knowledge around the midpoint of interpolation. The parallel scenario for shared knowledge remains relatively flat.

\input{gfx/results_table_acc_reverse}

\newpage
\section{Extended Discussion}
\label{app:extended_discussion}
As discussed in Section~\ref{sec:discussion} of the main text, our experiments investigating the linear merging of models highlight a distinct impact on shared versus task-specific knowledge, particularly when merging different checkpoints of an incrementally trained model. This extended discussion elaborates on these core findings, delving deeper into the interactions observed, the nuances between incremental and parallel adaptation scenarios, the broader implications for applying merging in CL, and considerations regarding when this approach might be most suitable. We begin by re-examining the different behavior of shared and specific knowledge components.

\textbf{Fate of Shared and Task-Specific Knowledge.}
A consistent observation across our experiments is the divergent behavior of shared versus task-specific knowledge under linear interpolation. Shared knowledge, whether established during pre-training or by specific features like a common visual cue learned by both endpoint models, is largely preserved and can even be enhanced by interpolation (Figure~\ref{fig:panel_d} and~\ref{fig:panel_e}). This aligns with findings where merging is thought to consolidate commonalities by finding a more central solution within a shared loss basin, often facilitated by a common training history~\citep{izmailov2018averaging, wortsman2022robust}.
At the same time, the minor performance variations observed, particularly the slight dip in the parallel scenario, hints at a nuanced interaction. While functionally near-identical (both proficient at $T_{\text{Color}}$), their underlying parametric solutions for this relatively simple, cue-specific skill might still diverge. This observation links to early findings on linear mode connectivity~\citep{frankle2018lottery}, where the extent of shared training iterations influences the ease of finding low-loss paths between solutions. In particular, subsequent independent fine-tuning that results in distinct features ($T_{\text{Color}}$ cue) can lead to slightly different local minima, ultimately resulting in (minor) barriers upon linear interpolation. Depending on the characteristics of the newly learned features, initialization in a common basin is no guarantee for future merging compatibility.
In sharp contrast, \textit{unshared task-specific knowledge} (e.g.,~when interpolating between a $T_{\text{Color}}$-specialized model and a $T_{\text{Noise}}$-specialized model) rapidly degraded upon interpolation. This highlights that distinct parameter adaptations optimized for unrelated specific cues are easily disrupted by averaging. This differential effect on shared versus unshared components aligns with observations made by \citet{zaman-etal-2024-fuse} regarding knowledge dynamics in language models.

\textbf{Benefits of Merging Sequentially Adapted Models.}
Another observation is the superior performance of merging models adapted in the \textit{incremental} (naive CL) scenario compared to the \textit{parallel} scenario, particularly for shared knowledge (Figure~\ref{fig:panel_e}, CIFAR-100 panel). When interpolating between sequentially trained states, the performance on the general CIFAR-100 task consistently surpassed that of merging parallel-trained endpoints, often exceeding even the original pre-trained model's accuracy. This suggests that the process of sequential adaptation, even without explicit CL mechanisms, moves the models along trajectories that are more amenable to beneficial merging. The shared trajectory of adaptation across tasks in the incremental setup could lead to more compatible or aligned representations of features, which are then effectively consolidated by interpolation. Similar results are presented by \citet{marouf2024weighted}. This finding is particularly relevant for CL, as it indicates that merging states from a single, continually evolving model might be more advantageous than merging independently trained specialists.

\textbf{Implications for Continual Learning.}
The observed degradation of unshared task-specific knowledge when merging models presents a complex consideration for CL. The fact that specific learned information can be lost during merging introduces a critical question of `utility'. If such specific adaptations represent undesirable artifacts like reliance on various biases or spurious correlations, as framed by \citet{zaman-etal-2024-fuse}, their erosion could be beneficial. However, in many CL contexts, the unique knowledge acquired for distinct past tasks is precisely what needs to be preserved, especially as retraining on all historical data is often infeasible. In these cases, naive merging would exacerbate catastrophic forgetting. This underscores the challenge of determining whether specific knowledge components are valuable or detrimental, a decision crucial for applying merging, but also other CL mechanism, effectively.

\textbf{When Might Merging Be Suitable for CL?}
Our results suggest that naive linear interpolation might be most promising for CL scenarios aiming to consolidate general, shared capabilities, or where tasks are sufficiently similar that their specific solutions are not too disconnected. It appears less suited for maintaining strong performance on diverse, specialized tasks if unique knowledge components are crucial. However, several factors could modify this outlook:
\textit{Task Similarity and Relatedness:} Merging models from closely related tasks may lead to less degradation. While one might question the gains if tasks are very similar, merging could still offer robustness by averaging out optimization noise or consolidating compatible specializations~\citep{izmailov2018averaging, ilharco2022patching}.
\textit{Model Architecture and Scale:}  Our study used a variant of ResNet-18. Much recent merging success has been on very large models. As works like \citet{ramasesh2021effect, ilharco2022patching} suggest, the effectiveness of patching or merging can improve with scale, possibly due to greater parameter capacity allowing for more disentangled~\citep{ortiz2023task} or robust encoding of specific knowledge. 
\textit{Merging Techniques:} 
Simple linear interpolation is a baseline. More advanced methods, including Fisher-weighted averaging~\citep{matena2022merging} or interference-resolving techniques~\citep{yadav2023ties, marczak2024magmax}, might better preserve specific knowledge. However, such methods primarily re-balance existing knowledge components; their utility still depends on the desirability of preserving those specific components.
\textit{Nature of Endpoint Models:} 
The training regimen of the endpoint models is critical. Recent work in continual pre-training suggests that specific optimization strategies for endpoints, such as using exponential moving averages, can significantly improve their suitability for subsequent merging and overall CL performance~\citep{udandarao2024practitioner, marouf2024weighted}. This points towards co-designing training and merging strategies.

Operationally, linear merging is often applied post-hoc and is computationally inexpensive compared to many online CL methods that modify the training process itself. While this offers flexibility, the success of merging implicitly still relies on the initial training trajectories leading to compatible solutions in parameter space, a condition often facilitated by shared pre-training or task similarity~\citep{frankle2018lottery}.

\end{document}

%% file: math_commands.tex
\usepackage{amsmath,amsfonts,bm}

\def\eqref#1{equation~\ref{#1}}

\def\1{\bm{1}}

\DeclareMathAlphabet{\mathsfit}{\encodingdefault}{\sfdefault}{m}{sl}
\SetMathAlphabet{\mathsfit}{bold}{\encodingdefault}{\sfdefault}{bx}{n}



%% file: gfx/results_table_acc_V2.tex
\begin{table}[htbp!]
\centering
\caption{\textbf{Numerical Results of the Experiments in the Main Text.} Displayed for each experiment is the classification accuracy (in \%) on the test set, reported as the mean $\pm$ standard error across three runs with different random seeds.}
\label{tab:accuracy}
\resizebox{\textwidth}{!}{%
\begin{tabular}{lllccccccccccc}
\toprule
\multicolumn{3}{c}{} & \multicolumn{11}{c}{Interpolation Coefficient $\alpha$} \\
\cmidrule(lr){4-14}
Training & Evaluation & Scenario & 0.0 & 0.1 & 0.2 & 0.3 & 0.4 & 0.5 & 0.6 & 0.7 & 0.8 & 0.9 & 1.0 \\
\midrule
\multirow{6}{*}{$T_{\text{Color}}$ $\rightarrow$ $T_{\text{Noise}}$} & \multirow{2}{*}{$T_{\text{Color}}$} & \textit{Incr.} & $98.58 \pm 0.05$ & $97.22 \pm 0.07$ & $94.85 \pm 0.13$ & $90.89 \pm 0.29$ & $85.22 \pm 0.29$ & $79.74 \pm 0.22$ & $74.60 \pm 0.20$ & $71.08 \pm 0.20$ & $68.50 \pm 0.19$ & $66.46 \pm 0.15$ & $64.97 \pm 0.24$ \\
& & \textit{Para.} & $98.57 \pm 0.05$ & $96.28 \pm 0.10$ & $90.25 \pm 0.37$ & $78.96 \pm 0.15$ & $69.56 \pm 0.23$ & $64.96 \pm 0.27$ & $62.69 \pm 0.25$ & $61.61 \pm 0.22$ & $60.92 \pm 0.20$ & $60.29 \pm 0.19$ & $59.43 \pm 0.26$ \\
\cmidrule(lr){2-14}
& \multirow{2}{*}{$T_{\text{Noise}}$} & \textit{Incr.} & $60.05 \pm 0.17$ & $60.70 \pm 0.15$ & $61.11 \pm 0.20$ & $61.83 \pm 0.10$ & $62.59 \pm 0.06$ & $64.19 \pm 0.11$ & $67.27 \pm 0.36$ & $73.63 \pm 0.75$ & $84.70 \pm 0.62$ & $95.45 \pm 0.26$ & $99.26 \pm 0.05$ \\
& & \textit{Para.} & $60.05 \pm 0.16$ & $60.16 \pm 0.18$ & $60.12 \pm 0.08$ & $60.20 \pm 0.03$ & $60.45 \pm 0.09$ & $60.88 \pm 0.03$ & $62.27 \pm 0.15$ & $65.46 \pm 0.40$ & $74.15 \pm 0.74$ & $92.78 \pm 0.21$ & $99.72 \pm 0.03$ \\
\cmidrule(lr){2-14}
& \multirow{2}{*}{No Cue} & \textit{Incr.} & $60.91 \pm 0.02$ & $61.53 \pm 0.06$ & $61.73 \pm 0.12$ & $61.99 \pm 0.11$ & $62.18 \pm 0.09$ & $62.40 \pm 0.09$ & $62.59 \pm 0.10$ & $62.69 \pm 0.02$ & $62.69 \pm 0.05$ & $62.43 \pm 0.15$ & $62.08 \pm 0.18$ \\
& & \textit{Para.} & $60.90 \pm 0.03$ & $60.98 \pm 0.10$ & $60.87 \pm 0.05$ & $60.85 \pm 0.10$ & $60.94 \pm 0.22$ & $60.93 \pm 0.09$ & $61.07 \pm 0.14$ & $61.17 \pm 0.17$ & $61.27 \pm 0.24$ & $61.23 \pm 0.18$ & $60.83 \pm 0.29$ \\
\midrule
\multirow{2}{*}{$T_{\text{Color}}$ $\rightarrow$ $T_{\text{Color}}$} & \multirow{2}{*}{$T_{\text{Color}}$} & \textit{Incr.} & $98.50 \pm 0.06$ & $98.59 \pm 0.04$ & $98.70 \pm 0.05$ & $98.76 \pm 0.05$ & $98.84 \pm 0.05$ & $98.91 \pm 0.05$ & $98.96 \pm 0.05$ & $99.01 \pm 0.03$ & $99.03 \pm 0.02$ & $99.07 \pm 0.02$ & $99.06 \pm 0.01$ \\
& & \textit{Para.} & $98.49 \pm 0.06$ & $98.25 \pm 0.05$ & $97.74 \pm 0.12$ & $97.24 \pm 0.23$ & $96.87 \pm 0.22$ & $96.71 \pm 0.25$ & $96.87 \pm 0.24$ & $97.22 \pm 0.16$ & $97.73 \pm 0.14$ & $98.17 \pm 0.06$ & $98.45 \pm 0.04$ \\
\bottomrule
\end{tabular}%
}
\end{table}

%% file: gfx/chunking_result_table.tex
\begin{table}[htbp!]
\centering
\caption{\textbf{Numerical Results for the ``Chunking'' Setup.} Displayed for each experiment is the test accuracy (in \%), reported as the mean $\pm$ standard error across three runs with different random seeds.}
\label{app:tab_chunking_accuracy}
\resizebox{\textwidth}{!}{%
\begin{tabular}{lllccccccccccc}
\toprule
\multicolumn{3}{c}{}{} & \multicolumn{11}{c}{Interpolation Coefficient $\alpha$} \\
\cmidrule(lr){4-14}
Training & Evaluation & Scenario & 0.0 & 0.1 & 0.2 & 0.3 & 0.4 & 0.5 & 0.6 & 0.7 & 0.8 & 0.9 & 1.0 \\
\midrule
\multirow{6}{*}{$T_{\text{Color}}$ $\rightarrow$ $T_{\text{Noise}}$} &  \multirow{2}{*}{$T_{\text{Color}}$ Cue} & \textit{Incr.} & $90.62 \pm 0.38$ & $89.43 \pm 0.40$ & $86.98 \pm 0.36$ & $83.55 \pm 0.40$ & $79.41 \pm 0.56$ & $75.00 \pm 0.58$ & $70.64 \pm 0.70$ & $66.29 \pm 0.78$ & $63.63 \pm 0.73$ & $55.76 \pm 0.71$ & $53.09 \pm 1.04$ \\
& & \textit{Para.} & $90.95 \pm 0.40$ & $87.93 \pm 0.30$ & $81.71 \pm 0.22$ & $72.95 \pm 0.55$ & $62.89 \pm 0.67$ & $54.67 \pm 0.55$ & $48.58 \pm 0.49$ & $46.12 \pm 0.44$ & $43.71 \pm 0.41$ & $41.35 \pm 0.41$ & $39.42 \pm 0.61$ \\
\cmidrule(lr){2-14}
& \multirow{2}{*}{$T_{\text{Noise}}$ Cue} & \textit{Incr.} & $41.62 \pm 0.63$ & $43.08 \pm 0.61$ & $44.49 \pm 0.64$ & $45.69 \pm 0.60$ & $46.92 \pm 0.61$ & $48.60 \pm 0.61$ & $52.76 \pm 0.62$ & $60.16 \pm 0.64$ & $75.12 \pm 0.94$ & $92.76 \pm 0.23$ & $99.12 \pm 0.05$ \\
& & \textit{Para.} & $41.62 \pm 0.63$ & $42.48 \pm 0.61$ & $42.82 \pm 0.64$ & $43.20 \pm 0.60$ & $43.60 \pm 0.61$ & $45.25 \pm 0.61$ & $49.08 \pm 0.62$ & $56.75 \pm 0.64$ & $74.12 \pm 0.94$ & $92.76 \pm 0.23$ & $99.12 \pm 0.05$ \\
\cmidrule(lr){2-14}
& \multirow{2}{*}{CIFAR-100} & \textit{Incr.} & $42.29 \pm 0.61$ & $43.60 \pm 0.52$ & $44.49 \pm 0.48$ & $45.34 \pm 0.47$ & $46.12 \pm 0.46$ & $46.53 \pm 0.68$ & $46.98 \pm 0.61$ & $47.02 \pm 0.48$ & $46.87 \pm 0.49$ & $46.15 \pm 0.55$ & $45.64 \pm 0.59$ \\
& & \textit{Para.} & $42.97 \pm 0.56$ & $43.18 \pm 0.50$ & $43.23 \pm 0.52$ & $43.29 \pm 0.47$ & $43.33 \pm 0.48$ & $43.15 \pm 0.32$ & $43.59 \pm 0.31$ & $43.66 \pm 0.31$ & $43.59 \pm 0.30$ & $43.02 \pm 0.30$ & $41.62 \pm 0.40$ \\
\midrule
\multirow{2}{*}{$T_{\text{Color}}$ $\rightarrow$ $T_{\text{Color}}$} &  \multirow{2}{*}{$T_{\text{Color}}$ Cue} & \textit{Incr.} & $90.62 \pm 0.38$ & $91.39 \pm 0.39$ & $91.93 \pm 0.35$ & $92.53 \pm 0.40$ & $93.07 \pm 0.40$ & $93.60 \pm 0.40$ & $93.94 \pm 0.40$ & $94.26 \pm 0.40$ & $94.56 \pm 0.40$ & $94.72 \pm 0.40$ & $94.73 \pm 0.40$ \\
& & \textit{Para.} & $90.81 \pm 0.27$ & $90.87 \pm 0.28$ & $90.83 \pm 0.22$ & $90.66 \pm 0.17$ & $90.39 \pm 0.16$ & $90.24 \pm 0.15$ & $90.36 \pm 0.17$ & $90.48 \pm 0.17$ & $90.68 \pm 0.16$ & $90.82 \pm 0.13$ & $90.79 \pm 0.04$ \\
\bottomrule
\end{tabular}%
}
\end{table}

%% file: gfx/results_table_acc_reverse.tex
\begin{table}
\centering
\caption{\textbf{Numerical Results for the Reverse Cue Order Experiments.} Shown for each experiment is the classification accuracy (in \%) on the test set, reported as mean $\pm$ standard error across three runs with different random seeds.}
\label{app:tab_cues_reverse_accuracy}
\resizebox{\textwidth}{!}{%
\begin{tabular}{lllccccccccccc}
\toprule
\multicolumn{3}{c}{} & \multicolumn{11}{c}{Alpha} \\
\cmidrule(lr){4-14}
Training & Evaluation & Scenario & 0.0 & 0.1 & 0.2 & 0.3 & 0.4 & 0.5 & 0.6 & 0.7 & 0.8 & 0.9 & 1.0 \\
\midrule
\multirow{6}{*}{$T_{\text{Noise}}$ $\rightarrow$ $T_{\text{Color}}$} & \multirow{2}{*}{Noise} & \textit{Incr.} & 99.76 $\pm$ 0.04 & 97.11 $\pm$ 0.33 & 87.44 $\pm$ 1.00 & 76.17 $\pm$ 0.99 & 69.57 $\pm$ 0.66 & 66.34 $\pm$ 0.37 & 64.68 $\pm$ 0.15 & 63.83 $\pm$ 0.04 & 63.36 $\pm$ 0.05 & 62.87 $\pm$ 0.18 & 62.39 $\pm$ 0.20 \\
& & \textit{Para.} & 99.76 $\pm$ 0.04 & 93.70 $\pm$ 0.23 & 75.86 $\pm$ 0.14 & 66.39 $\pm$ 0.17 & 62.76 $\pm$ 0.06 & 61.31 $\pm$ 0.04 & 60.71 $\pm$ 0.06 & 60.51 $\pm$ 0.06 & 60.51 $\pm$ 0.12 & 60.48 $\pm$ 0.09 & 60.29 $\pm$ 0.12 \\
\cmidrule(lr){2-14}
& \multirow{2}{*}{Color} & \textit{Incr.} & 59.71 $\pm$ 0.12 & 60.70 $\pm$ 0.13 & 61.84 $\pm$ 0.02 & 63.03 $\pm$ 0.17 & 65.05 $\pm$ 0.23 & 68.59 $\pm$ 0.24 & 74.77 $\pm$ 0.28 & 83.52 $\pm$ 0.22 & 90.32 $\pm$ 0.05 & 94.25 $\pm$ 0.14 & 96.21 $\pm$ 0.13 \\
& & \textit{Para.} & 59.71 $\pm$ 0.12 & 60.23 $\pm$ 0.10 & 60.60 $\pm$ 0.06 & 60.94 $\pm$ 0.04 & 61.73 $\pm$ 0.03 & 63.36 $\pm$ 0.08 & 67.26 $\pm$ 0.28 & 75.35 $\pm$ 0.49 & 87.22 $\pm$ 0.30 & 94.63 $\pm$ 0.26 & 97.59 $\pm$ 0.10 \\
\cmidrule(lr){2-14}
& \multirow{2}{*}{CIFAR-100} & \textit{Incr.} & 61.07 $\pm$ 0.16 & 61.66 $\pm$ 0.13 & 61.98 $\pm$ 0.13 & 62.38 $\pm$ 0.15 & 62.68 $\pm$ 0.18 & 62.75 $\pm$ 0.16 & 62.82 $\pm$ 0.16 & 62.99 $\pm$ 0.20 & 63.12 $\pm$ 0.17 & 62.98 $\pm$ 0.11 & 62.81 $\pm$ 0.13 \\
& & \textit{Para.} & 61.07 $\pm$ 0.16 & 61.35 $\pm$ 0.17 & 61.45 $\pm$ 0.20 & 61.43 $\pm$ 0.20 & 61.15 $\pm$ 0.13 & 61.02 $\pm$ 0.12 & 60.88 $\pm$ 0.08 & 60.94 $\pm$ 0.04 & 60.99 $\pm$ 0.06 & 61.14 $\pm$ 0.11 & 60.85 $\pm$ 0.13 \\
\bottomrule
\end{tabular}
}
\end{table}